\documentclass[conference]{IEEEtran}
\IEEEoverridecommandlockouts
\usepackage{cite}
\usepackage{amsmath,amssymb,amsfonts}
\usepackage{algorithm}
\usepackage{algorithmic}
\usepackage{graphicx}
\usepackage{microtype}
\usepackage{subfigure}
\usepackage{textcomp}
\usepackage{xcolor}
\usepackage{subfigure}
\usepackage{booktabs} % for professional tables
\usepackage{amsmath} % to use \operatorname*
\usepackage{amsthm}
\usepackage[outdir=./]{epstopdf}
\usepackage{url}

\def\BibTeX{{\rm B\kern-.05em{\sc i\kern-.025em b}\kern-.08em
    T\kern-.1667em\lower.7ex\hbox{E}\kern-.125emX}}

\newcommand{\Softneuro}{SoftNeuro}
\newcommand{\softneuro}{SoftNeuro}
\newcommand{\fastest}{\operatorname*{fastest}}

\newcommand{\argmin}{\mathop{\rm arg~min}\limits}

\theoremstyle{definition}\newtheorem{definition}{Definition}
\theoremstyle{plain}\newtheorem{proposition}{Proposition}

\begin{document}

\title{SoftNeuro: Fast Deep Inference using Multi-platform Optimization\\}
\author{\IEEEauthorblockN{Masaki Hilaga, Yasuhiro Kuroda, Hitoshi Matsuo,  Tatsuya Kawaguchi, \\
Gabriel Ogawa, Hiroshi Miyake and Yusuke Kozawa}
\IEEEauthorblockA{
\textit{Morpho, Inc.}\\
Tokyo, Japan \\
(m-hilaga, y-kuroda, h-matsuo, t-kawaguchi, g-ogawa, h-miyake, y-kozawa)@morphoinc.com}
}

\maketitle

\begin{abstract}

% 深層学習は様々な分野で実用化されつつあり、その高い計算コストをいかに抑
% えるかというのは, カーボンオフセットのためにも, 特に推論時において重要
% な課題となる.

% この問題を解決するために, 我々は新しい推論フレームワークを
% 提案する. そこではあるレイヤーに対して様々なアルゴリズムのルーチンをプ
% ロファイルし, 最速のものを選択することでパフォーマンスを最適化する.

% 我々は探索範囲を削減した上での動的計画法で最速パスを求める新しいアルゴリ
% ズムも提案し、これにより非常に効率的なパジョーマンス最適化を実現する.

  % Deep learning has seen increased practical use in a wide range of fields.
  Faster inference of deep learning models is highly demanded on edge devices and even servers,
  for both financial and environmental reasons.  To address this issue,
  we propose \textit{SoftNeuro},  a novel, high-performance inference framework with efficient performance
  tuning.  The key idea is to separate algorithmic routines from network layers.  Our
  framework maximizes the inference performance by profiling various routines for each
  layer and selecting the fastest path.  To efficiently find the best path, we propose a
  routine-selection algorithm based on dynamic programming.  Experiments show that the proposed
  framework achieves both fast inference and efficient tuning.

\end{abstract}


\section{Introduction}
\label{introduction}

% Deep neural network (DNN) はcomputer visionや音声認識、自然言語処理を含む様々なドメインで著しい精度の向上を見せており、実世界での応用も広がっている。
% computer vision領域においてはconvolutional neural network (CNN)を応用したmodelが物体認識や物体検出においてstate-of-the-artの結果を出しているし、自然言語処理分野ではTransformer機構を用いたテキスト生成や翻訳が注目されている。

Deep neural networks (DNNs) have shown remarkable accuracy improvement in various
domains, including computer vision, speech recognition, and natural language processing.
In the field of computer vision, models based on convolutional neural networks (CNNs)
have produced state-of-the-art results in object recognition, detection, and
segmentation~\cite{liu:ijcv20, zeng:arxiv21, minaee:arxiv20}.  Text generation and
translation using the transformer mechanism are also attracting attention in natural language processing~\cite{brown:arxiv20}.
More and more real-world applications have come to take advantage of these technologies.

% 大量のデータを用いて、DNN modelを学習する際にはよく知られているpytorchやtensorflow, MXNetといったフレームワークを用いることができる。
% ただし、これらのフレームワークは、GPU等リソースが十分にある環境に対して巧妙に最適化されたライブラリに依存している。
% そのため、例えばスマートフォンやAR/VRデバイス、産業用機器など、計算リソースとメモリの限られたエッジデバイス上での実応用の場面ではこれらのフレームワークが選択されることは少ない。

DNN models can be trained by well-known frameworks such as PyTorch \cite{pytorch2019},
TensorFlow \cite{tensorflow2016}, and MXNet \cite{mxnet2015}.  These frameworks achieve
efficient training by taking advantage of well-optimized libraries.
However, those libraries are only tuned for a narrow range of environments with sufficient resources such as GPUs.
These frameworks will not run as well on edge devices, such as smartphones, AR/VR devices, and industrial equipment, or
even less powerful servers, due to limited computational resources and memory. Improving computational capability is often
prohibitively expensive or impractical, be it from material costs, thermal issues, or even power consumption related costs or
environment-friendliness. Inference optimization is thus an important task; with sufficiently fast inference, those devices can be used practically,
without additional computation capability.

% エッジデバイス上での推論へのニーズは高まっており、自動微分を伴う学習の機能を持たない、推論のためのライブラリがいくつもリリースされている。
% Tensorflow Lite, Pytorch Mobileは代表的な学習フレームワークで学習した結果をエッジデバイスにおいて効率的に実行するためのツールセットである。
% これらは各デバイス向けにチューニングされた関数を用い、また量子化やpruningを含むモデルの最適化を行う事によりデバイス上で高速な推論を行うことができる。
% また、ハードウェアベンダーによってOpenVINOやTensorRT, SNPEなどといった推論エンジンがリリースされている。
% これらは特定のHWリソースを最大限活用する工夫がHWベンダーによりなされているという意味で、使用するHWが決まっている場合には有効である。
% ただしこれらの推論用ライブラリも各種演算の最適化は大変に巧妙であり高速な関数が用意されているが、グラフレベルの最適化については十分に実現されているとは言えない。

With the above motivation, several inference focused libraries have been released for fast inference on edge devices.
TensorFlow Lite~\cite{tfl} and PyTorch Mobile~\cite{ptm} are representative examples of such
toolsets, and efficiently perform inference of models trained by the aforementioned
frameworks.  These toolsets maximize performance by carefully tuning
operations for each device and optimizing models with well-known techniques such as
quantization and pruning.  Meanwhile, hardware vendors have also released inference
engines such as OpenVINO \cite{openvino2019}, TensorRT \cite{trt}, and SNPE \cite{snp}.
These engines would be most effective when the deployment target hardware is predetermined,
as each vendor optimizes for their own hardware.
Furthermore, although these inference libraries are successful in optimizing individual operations of DNNs, they do
not provide sufficient optimization at the graph level.

% 一方、TVMは演算のスケジューリングを適切に行うことで様々なデバイス上での高速な推論を可能としており、近年AutoTVMにより機械学習ベースでのパラメータチューニングも提案されているが、〜〜〜という問題点が残っている。

Another framework, named TVM \cite{tvm}, optimizes inference on various devices by
appropriately scheduling operations.  AutoTVM \cite{autotvm2018} has also been proposed to
tune parameters using machine learning.  However, the search space is limited to a small
part of the many possible scheduling combinations, often making optimization insufficient.
Moreover, its tuning is quite time-consuming, as shown in our experiments.
% citation needed?

In this paper, we address the aforementioned issues by introducing a framework that
enables efficient tuning over large search spaces at the graph level.  The framework
performs fast inference of deep learning models trained by the learning frameworks on a
wide range of platforms, including CPUs, GPUs, and DSPs.  The main idea behind efficient
tuning is separating algorithmic routines from network layers of DNN models
(Section~\ref{basic_architecture}).
% Layers are abstract structures that only define a
% computation type.  On the other hand, routines are concrete implementations of layer computations,
% allowing for the application of multiple algorithms per computation type, which is useful because
% selecting a suitable algorithm heavily depends on various factors such as algorithm parameters, target
% environments, and data characteristics.

Our framework maximizes inference performance by using actual execution measurements on target devices.
This optimization takes into account multiple
aspects: target deployment devices (e.g., CPUs, GPUs, and DSPs), data types (e.g.,
float32, float16, and qint8), data layout (e.g., channels-last and channels-first), and
algorithms used (e.g., direct, Winograd, and sparse).  Since a naive algorithm to find the
best routines at the graph level may hit combinatorial explosions, we propose an
optimization algorithm based on dynamic programming (Section~\ref{optimization}).

Experiments in Section~\ref{experiments} show that our framework is as much as 3 times faster than other
inference frameworks on a Galaxy S8 smartphone equipped with a Snapdragon 835 SOC.
In addition, the tuning of our framework can be 500 times faster than that of
TVM for the VGG16 model~\cite{simonyan2015very}.

\section{Related Work}
\label{related_work}

We review related work from the viewpoint of accelerating the convolution operation
because it occupies considerable inference time, making its efficient calculation
essential~\cite{liu:atc19}.
% A typical deep learning model consists of a number of convolutional layers and other
% layers.
% A commonly used 2D convolutional layer is represented as follows
% \begin{equation}
%   O_{y,x}[c'] = \sum_{ky,kx,c}{I_{y+ky,x+kx}[c]W_{ky,kx}[c, c']},
% \end{equation}
% where $I_{y,x}[c]$ is the input, $O_{y,x}[c']$ is the output, and $W_{ky,kx}[c, c']$ is
% the kernel.
There are two important factors towards accelerating convolutions' computation~\cite{adams2019optimizehalide}: 
(1) how computations are performed (\emph{algorithms}) and
(2) how computations are divided and in which order they are
executed (\emph{scheduling}).  Scheduling optimization is described in
Section~\ref{sec:scheduling}, and algorithmic improvements are summarized in
Sections~\ref{sec:winograd} and \ref{sec:quantization}.

\subsection{Automatic Scheduling}
\label{sec:scheduling}

Most inference frameworks simply select computation methods based on some
heuristics~\cite{tfl, mnn}.  Other frameworks such as Halide
\cite{halide} and TVM \cite{tvm} optimize scheduling independently of employed algorithms.
Halide \cite{halide} uses a genetic algorithm to search for an approximate optimal schedule solution.
In \cite{adams2019optimizehalide},  improvemrnts to Halide are proposed by enabling larger
search spaces of schedules and employing a tree search algorithm based on beam
search.  TVM \cite{tvm} separates hardware intrinsics from algorithms and
scheduling, supporting unseen devices.  To optimize scheduling, they
predict running times of schedules with machine learning models based on XGBoost
\cite{xgboost} by using features such as access to various memories.

% Issues and our framework
While the above methods can optimize model inference as a whole, the number of scheduling
combinations is quite large and expensive to search, so limiting the scheduling space and improving search 
efficiency are still important issues.  We address these issues by introducing
the concepts of routines and layers with an efficient optimization algorithm based on
profiling.

\subsection{Winograd Algorithm}
\label{sec:winograd}

Winograd's \textit{minimal filtering}~\cite{winograd} was originally proposed as a method to
minimize the number of multiplications during convolution in the context of digital signal
processing.  Nevertheless, it can also be extended to convolution in CNNs~\cite{lavin2016},
and is implemented by various inference frameworks.
The Winograd algorithm simultaneously calculates convolution over tiles for a specific
tile size, but the optimal tile size needs to be properly chosen for the best performance.
Some inference frameworks estimate the appropriate tile size based on input characteristics
such as data sizes~\cite{mnn}.  We address this issue by measuring the performance with
multiple tile sizes and selecting the best setting.

\subsection{Quantization}
\label{sec:quantization}

% DNNのモデルサイズの削減と計算コスト削減のためにweightやactivationの量子化もよく使用される手法である。
% 量子化は、通常浮動小数点数として扱われる値を整数と少数のパラメータで表現する手法のことで次のような式で表される。
% q = offset + round(x / scale)
% ここでoffsetは整数、scaleは実数である。
% これらの係数の決定方式には複数あり、offsetが0の場合はsymmetricな量子化と呼ばれ、そうではない場合はasymmetricな量子化と呼ばれる。
% また、テンソル毎にquantizeを行うか、channel毎にquantizeを行うかによって精度とモデルサイズのトレードオフが生じる。
% また学習ずみモデルを量子化するpost-quantizationだけでなく、tensorflowなど一部の学習フレームワークでは精度を担保するためにquantization-aware trainingの仕組みも導入されている。
% さらにはweight, activationをa few bitsまで低減したTernary weight networks, Binary Neural Networks, XNOR-net なども提案されているが現在の手法は精度へのインパクトが大きい。

Quantization of weights and activations is another common technique to reduce the model
size and computational cost of DNNs~\cite{sze:procieee17}.  Quantization expresses values (e.g., 32-bit floating point number) 
at a lower-precision (e.g., 8-bit) integer and a small number of parameters.
% \begin{equation}
% q = c + {\rm round}(x / s)
% \end{equation}
% Where $c$ is an integer offset, and $s$ is a scale.  There are several ways to determine
% these coefficients: if the offset is zero, it is called symmetric quantization; otherwise,
% it is asymmetric.  There is also a trade-off between accuracy and model size depending on
% whether quantization is done per tensor or per channel. % is there? why? citation needed
% In addition to post-training quantization,
Some training frameworks such as TensorFlow~\cite{tensorflow2016} have introduced a
quantization-aware training mechanism to improve
accuracy~\cite{krishnamoorthi2018quantizing}.  In addition, inference engines such as
TensorFlow Lite~\cite{tfl}, PyTorch Mobile~\cite{ptm}, and NVIDIA TensorRT~\cite{trt}
support 8-bit integer quantization.  \Softneuro{} also implements quantization-based
routines to accelerate convolutions, as well as other computations.

% Moreover, Ternary weight networks \cite{Li2016TernaryWN}, Binary Neural Networks
% \cite{Courbariaux2016BinarizedNN}, and XNOR-net \cite{Rastegari2016XNORNetIC}, which
% reduce the weight and activation to a few bits, have also been proposed.

\section{Framework Features}
\label{basic_architecture}

This section gives an overview of \softneuro{} features: how deep learning models are
deployed in \softneuro{} (Section~\ref{model_deployment}), how it represents
deep learning models (Section~\ref{model_architectures}), and how it optimizes model
execution on various platforms (Section~\ref{routines_and_profiling}). It also introduces
\emph{childnets}, structures that facilitate the implementation of additional,
high-performance routines by reusing existing ones (Section~\ref{childnet}).

% 基本的アイデア,
% つまり各レイヤに対して異なる手法のルーチンを,
% 実際の実行環境で実行しその処理速度を測定,
% その上で適切なルーチンを各レイヤに割り当てるという話.

% Import, profile, optimize, run に至るまでの一連の処理の流れの話.
\subsection{Model Deployment}
\label{model_deployment}

% Insert a figure?

% Overview
A DNN model can be deployed with \softneuro{} in three major steps: (1) Import a DNN
model trained by popular deep learning frameworks, converting it into \softneuro's
\texttt{dnn} format. (2) Tune the model to maximize the inference speed on the target
platform. (3) Perform inference using the tuned model.  A brief description of the
main components (importers, profiler, and optimizer) needed for deployment follows.

% Importer
Importers are responsible for two tasks: (1) Convert a given DNN model into \softneuro's
format, and (2) Optimize the computational graph of the model.  Importers currently support conversion from
TensorFlow~\cite{tensorflow2016} and ONNX\footnote{\url{https://onnx.ai/}} (which in turn
supports Caffe2~\cite{caffe2}, Chainer~\cite{chainer}, Microsoft Cognitive
Toolkit~\cite{cntk}, MXNet~\cite{mxnet2015}, PyTorch~\cite{pytorch2019}, and
PaddlePaddle~\cite{paddlepaddle}).  When converting a model, importers automatically
optimize the model's computational graph.  For instance, ReLU activation and batch
normalization layers are fused into their previous layers, reducing computational
cost without affecting accuracy.

% Profiler & Optimizer
The profiler and optimizer tune models to run on a given target platform as fast as
possible.  To this end, \softneuro{} separately considers \emph{layers} and
\emph{routines}.  Layers are an abstract concept that define only what kind of computation
should be performed (e.g., ReLU and 2D convolution).  On the other hand, routines are
actual implementations of layer computations optimized for various hardware platforms
(e.g., Intel CPUs, NVIDIA GPUs) and data types (e.g., float32, qint8).  A \texttt{dnn} model
can be regarded as a graph of layers, and tuning is accomplished by selecting each layer's fastest
routine.  For tuning, the profiler measures the processing times of
available routines for each layer, and by using the profiling data, the optimizer selects
the best routines for layers.  Further details can be read in
Sections~\ref{model_architectures} and \ref{routines_and_profiling}.

% モデルアーキテクチャの話 (model, net, layer, souce layer, sink layer, ...).
\subsection{Model Representation}
\label{model_architectures}

% net overview
A \texttt{dnn} model consists of a \emph{net}, a directed acyclic graph (DAG) of
layers.  Nodes of the graph are layers, and edges represent computational dependencies
between layers.  A net handles input in the first layer, and the net output is obtained
from the last layer.  Other layers between input and output perform computations to
calculate the final output.

% Layers
A layer is an abstract structure that takes one or more $n$-dimensional tensors, or
\emph{blobs}, performs some computations, and outputs the resulting blob.  Every layer has a
\emph{type}, which defines what kind of computation should be performed.  For instance, a
layer of the \texttt{conv2} type takes an input blob, carries out 2D convolution with a
layer-specific kernel and bias, and outputs the convolution result.

% Layer parameters and weights
Layers contain two kinds of attributes, namely \emph{layer parameters} and \emph{weights}.
Layer parameters specify how to execute the layer type's computation, and weights are
trained parameters specific to the layer.
For instance, layer parameters of a \texttt{conv2} layer include dilations, strides, and padding,
while the weights would be the kernel and bias.

% Example?

% ルーチンとプロファイリングの話
% (routine, routine descriptor, routine parameters,
% ルーチンパラメの組み合わせでプロファイル).
\subsection{Routines and Profiling}
\label{routines_and_profiling}

% Routines and routine descriptors
Routines are concrete implementations of layers, and many implementations can exist in
\softneuro{} for each layer.
To discriminate routines, we introduce \emph{routine descriptors}.
A descriptor specifies the platform, algorithm, and data type each routine was made for.
For instance, the \texttt{cuda:float16/cudnn} routine for a \texttt{conv2}
layer uses the cuDNN library~\cite{cudnn} to perform 2D convolution in half-precision on
a compatible GPU.

% Routine schema and adapt layers
The part before the slash (\texttt{cuda:float16} in the example) is called \emph{routine
  schema}.  It defines a routine's platform and data type.  If schemas are different
between routines of two adjacent layers, the data format or data location must be
changed appropriately.  For instance, if a layer has a \texttt{cpu} routine and the next
layer has a \texttt{cuda} routine, data must be transferred to a GPU before computing
the second layer.  To this end, \softneuro{} automatically inserts an \emph{adapt} layer
between such two layers.  Adapt layers handle three kinds of jobs: (1) data transfer
between heterogeneous devices, (2) casting data types, and (3) changing data layout.

% Profiling
Given there are multiple routines, we need to select each layer's best routine when
deploying models.  However, each routine's performance heavily depends on its algorithm
and various characteristics of the target platform (e.g., cache structures).  Thus, it is
practically impossible to determine in advance the best combination of routines for all
possible platforms.  We address this issue by profiling available routines first, and then
selecting the best-performing routine.

% Routine parameters
To accommodate diverse characteristics of multiple devices, we add
\emph{routine parameters} to each routine.  Routine parameters specify values of internal
variables used in routines.  For instance, the \texttt{cpu/woc64\_avx} routine\footnote{A \texttt{conv2} routine that utilizes AVX for computation.}  has two routine parameters: cache size
and task granularity.  Each parameter is a list of integers, and the profiler considers
all of the possible combinations of the parameters, thereby enabling \softneuro{} to
select the best performing routines on many kinds of devices.

% Unit and integrated profiling
Actual profiling is divided into two steps: (1) \emph{unit profiling}, which 
profiles routines in a layer-by-layer fashion, and (2) \emph{integrated profiling}, which
considers all of a net's routines as a whole.  This division aims to
reduce the computational cost of integrated profiling. Ideally we would perform
integrated profiling from scratch, but the search space for a whole net can be
combinatorially large.  On the other hand, performing only unit profiling can be an alternative
strategy, but it is insufficient due to possible differing performance between unit
and integrated cases, mainly due to cache behavior. Therefore we opt for using
unit profiling to filter out unpromising routine patterns, reducing the combinations for
integrated profiling.
% Assumption: If a routine is good in unit profiling, then the routine is good also in
% integrated profiling.
Specifically, our current implementation selects routine patterns performing better in
unit profiling so that the number of patterns in integrated profiling is judiciously limited to a searchable
amount.

% Example?

\subsection{Childnet}
\label{childnet}

\begin{figure}[t]
  \centering
  \subfigure[\nolinebreak \mbox{A dense routine.}]{%
    \includegraphics[width=0.26\linewidth]{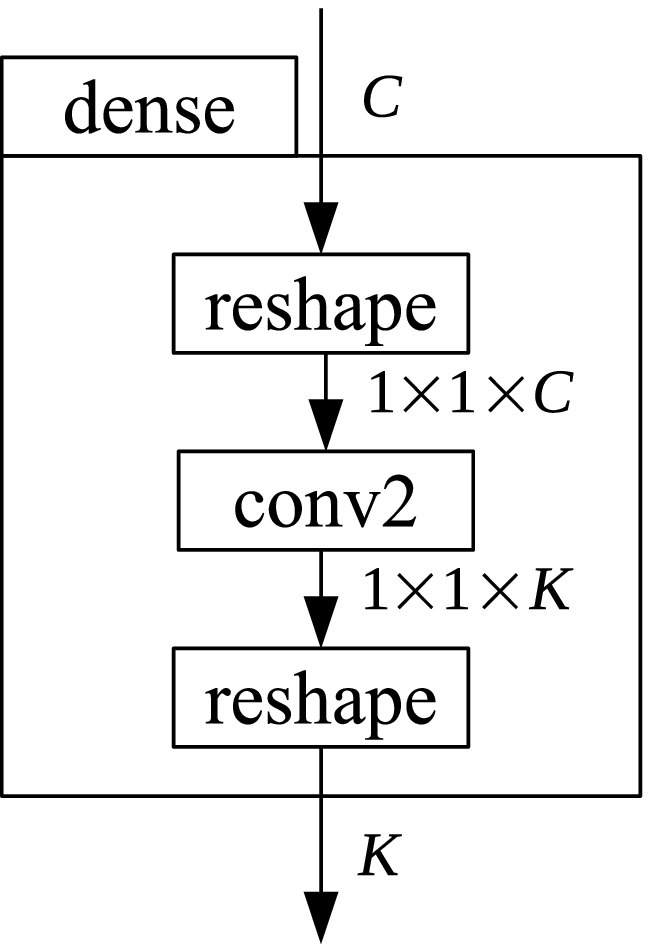}
    \label{childnet-dense}
  }
  \hspace{2em}
  \subfigure[\nolinebreak \mbox{A Winograd routine.}]{%
    \includegraphics[width=0.26\linewidth]{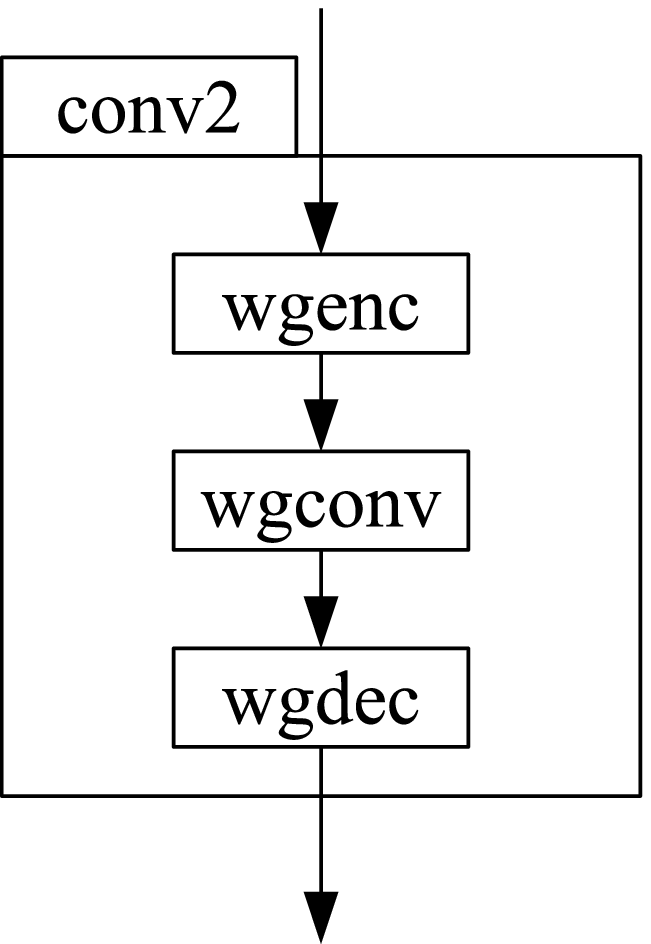}
    \label{childnet-winograd}
  }
  \caption{Routines implemented by childnets.}
  \label{childnet-examples}
\end{figure}

% Childnets
We introduce a mechanism called \emph{childnet}, which enables us to implement a new
routine by reusing existing ones, for the following two objectives: (1) to facilitate
the implementation of additional routines,  and (2) to allow for efficient profiling.  In the
following, we describe how childnets achieve the objectives by using
Figure~\ref{childnet-examples} as examples.

% First objective
In Figure~\ref{childnet-dense}, we demonstrate that childnets can easily implement a \textrm{dense} routine.
Three sub-layers compose the routine: \texttt{reshape}, \texttt{conv2}, and \texttt{reshape}.  The first
\texttt{reshape} layer changes the input shape from $C$ to $1 \!\times\! 1 \!\times\! C$.
The \texttt{conv2} layer uses the reshaped blob for 2D convolution with a kernel of shape
$1 \!\times\! 1 \!\times\! C \!\times\! K$.  The result of convolution is reshaped from
$1 \!\times\! 1 \!\times\! K$ to $K$ in the last layer.  In this way, a new, efficient
routine can be easily constructed by combining existing routines.

% Second objective
In addition, childnets allow for efficient profiling of specific routines by exploiting
the integrated-profiling mechanism.  An example of such routines is a Winograd routine for
2D convolution.  A naive implementation of the routine without childnet contains four
kinds of routine parameters.  In this case, profiling needs to be performed for all
combinations of routine parameters.  On the other hand, a Winograd routine can also be
implemented using the childnet mechanism as shown in Figure~\ref{childnet-winograd}.
This Winograd routine consists of three sub-layers: \texttt{wgenc}, \texttt{wgconv}, and
\texttt{wgdec}.  Each routine of these layers has only a single routine parameter, and the
Winograd routine itself also has a routine parameter.  In this case, we can reduce the
number of patterns to be profiled with the filtering mechanism of integrated
profiling, thereby achieving efficient profiling.

\section{Algorithm for Performance Optimization}
\label{optimization}

% Why is the algorithm necessary?
In this section, we describe the proposed algorithm for performance optimization of
routines using the profiling results.  If we could assume only a single routine
schema, it would be sufficient to use a simple algorithm that selects each layer's fastest
routine.  However, the difficulty arises when we handle multiple schemas
with adapt layers, because we also need to incur the cost of adaptation performed by the
adapt layers (e.g., data format conversion).  Such tuning scenarios are called
\emph{hybrid tuning} in this paper, and our algorithm for hybrid tuning is named
\emph{dynamic programming for routine selection}, or \emph{DPRS} for short.

% Section overview
In the following, we first give basic notations and definitions in
Section~\ref{definitions}.  Then Sections~\ref{sec:straight_net}--\ref{sec:mimo_net}
explain the algorithm from the simplest case to gradually complex cases.
Section~\ref{sec:dprs_algorithm} summarizes the algorithm workflow by describing its
pseudo-code.

\begin{figure*}[tb]
  \centering
  \includegraphics[width=0.7\linewidth]{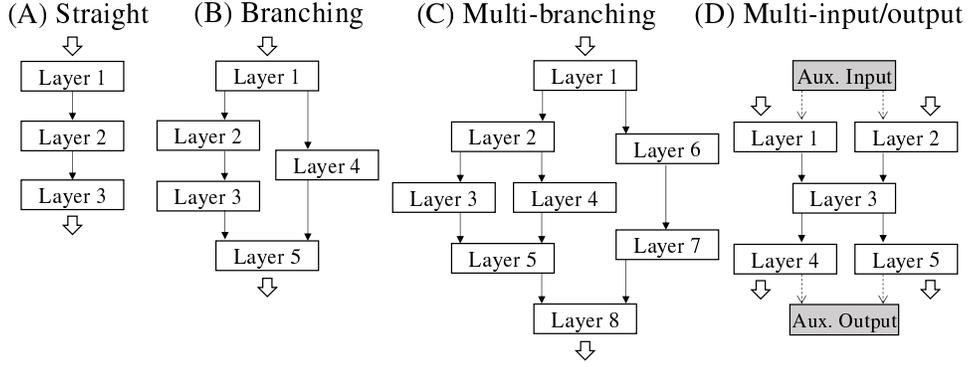}
  \caption{Network structure examples.}
  \label{fig:network_examples}
\end{figure*}

\subsection{Notations and Definitions}
\label{definitions}

Our algorithm's objective is to select the best routine for each layer, given a net, a set of routine
schemas, and profiling data.  To formulate the problem, we begin with a formal definition of nets.

% Net and routines
A net is a DAG with one input and one output (\emph{1-in-1-out net}) consisting of $n$
layers.  The layers are denoted as $L_1, \ldots, L_n$ in topological order, where $L_1$ is
the input layer and $L_n$ is the output layer.  Each layer $L_i$ has a set $R_i$ of
routines. A set $R_i(\lambda)$ is the subset of $R_i$ that contains the routines with
schema $\lambda$.  We define the fastest routine $F_i(\lambda)$ within $R_i(\lambda)$ as
\begin{equation}
  F_i(\lambda) \equiv\argmin_{r\in R_i(\lambda)} T(r),
\end{equation}
where $T(r)$ is the execution time of routine $r$ measured by profiling.

% Optimal routine path is the desired result
Given these notations, we provide the definition of \emph{routine path} as follows.

\begin{definition}[Routine path]
\label{def:routine_path}
A routine path $S$ is a possible set of routine combinations for a given net.  It consists
of each layer's routine as $S=\{r_i \in R_i \mid 1 \leq i \leq n \}$.  When the edge
$(L_i, L_j)$ exists in the net and their routines ($r_i$ and $r_j$) have different
schemas, the adapt routine $r^a_{i, j}$ is also included in the set $S$.
\end{definition}

Based on this definition, we define the \emph{optimal routine path}, which is the desired
result for hybrid tuning.

\begin{definition}[Optimal routine path]
\label{def:optimal_routine_path}
% 最適ルーチンパス$S^*$は、すべてのルーチンパスの中で全体の処理時間が最小のものを意味する。
% すなわち以下のような組み合わせ最適化の解である。
The optimal routine path $S^*$ is the path with the minimum overall processing time among all routine paths for a net.
In other words, it is the solution of the following combinatorial optimization:
\begin{equation}
\label{eq:optimal_routine_path}
S^* = \argmin_S T(S),
\end{equation}
% ただし$T(r)$はルーチンパスに対しては$T(S)=\sum_{r\in S}T(r)$が成り立つものとする。
where $T(S)=\sum_{r\in S}T(r)$ for routine path $S$.
\end{definition}

\subsection{Straight Net}
\label{sec:straight_net}

%図\ref{fig:network_examples}Aのような場合でルーチンディスクリプタ$\Lambda=\{\alpha,\beta\}$でハイブリッドチューニングする場合を考える。
% 各レイヤーの処理をいずれの型で行うかの選択肢があるので、$S^*$の探索スペースは$O(2^N)$である。
% しかし以下の動的計画法によってこれを$O(N)$に削減することができる。
% まず$\forall \lambda\in\Lambda, S_1(\lambda)=\{F_1(\lambda)\}$と初期化する。以降$i=2,\ndots, N$について、

We first consider the simplest case illustrated as the net A in Figure
\ref{fig:network_examples}, over the routine schemas $\Lambda=\{\alpha, \beta\}$.  In this
case, the size of search space for $S^*$ is $O(2^n)$ because we have two options (i.e.,
$\alpha$ and $\beta$ routines) for each of $n$ layers.  To reduce the computational
complexity to $O(n)$, we employ a simple dynamic-programming approach, formulated as the
following recurrence relation:
\begin{equation}
  \label{eq:recurrence_formula}
  \begin{array}{l}
    S_i(\alpha)=\{F_i(\alpha)\}\cup\fastest
    \left\{\begin{array}{l}
             S_{i-1}(\alpha), \\
             S_{i-1}(\beta) \cup \\
             ~~~\{F^a_{i-1,i}(\beta \rightarrow \alpha)\}
           \end{array},
    \right.
  \end{array}
\end{equation}
where $F^a_{i,j}(\beta \rightarrow \alpha)$ denotes the fastest adapt routine for layer
$i$ to layer $j$, and at $i = 1$, $S_1(\lambda) = \{F_1(\lambda)\}$ for
$\lambda \in \Lambda$.  The $\fastest$ construct is an operation that takes routine paths
and finds the fastest path.

The relation means that the optimal routine subpath at layer $i$ can be obtained by
selecting the fastest routine for layer $i$ and computing the optimal routine subpath at
layer $i - 1$.  The latter is either $S_{i-1}(\alpha)$ or $S_{i-1}(\beta)$ with an adapt
routine, because we need to include the adaptation cost when schemas change.  While the
above relation is for schema $\alpha$, a similar relation for schema $\beta$ also holds.
By using the relations, we can show the following proposition.

\begin{proposition}[Optimality of DP]
\label{prop:optimality_of_dp}
The routine path $\hat{S} = \fastest_{\lambda \in \Lambda}{S_n(\lambda)}$ is optimal.
\end{proposition}
%

%\begin{proof}
  % 証明：任意の$i$において、$S_i(\lambda)$は$L_1,\ndots,L_i$のルーチンサブパ
  % ス$S$で、$r_i\in
  % S$が$\lambda$に当てはまるものの中で最適であるということを、帰納法で示すことができる。帰
  % 納法は式(\ref{eq:recurrence_formula})から明らかである。
 \noindent{\textit{Proof.} It is shown by induction that for any $i$, $S_i(\lambda)$ is the optimal routine subpath
  of $L_1, \dots, L_i$, where routine $r_i$ has schema $\lambda$. This induction is clear
  from the equation \ref{eq:recurrence_formula}.}
%\end{proof}

\subsection{Branching Net}
%図\ref{fig:network_examples}Bのように分岐を含むネットワークをハイブリッドチューニングする場合について説明する。
% レイヤー５における最適ルーチンサブパス$S_5(\lambda_5)$は$S_3(\lambda_3)$と$S_4(\lambda_4)$とレイヤー３と５、レイヤー４と５を接続するアダプトルーチンの和集合で表すのが自然である
% そのためには$r_1\in S_3(\lambda_3)$と$r_1'\in S_4(\lambda_4)$が等しくなければいけない。このようなケースを扱うために$S_i(\lambda)$を拡張し、$S_i(\lambda\mid C)$を定義する。
% ここで、$C$はルーチンの集合で、拘束条件$C\subseteq S_i(\lambda\mid C)$を意味する。この拡張によってレイヤー５の最適ルーチンサブパスを以下の式により漸化的に求めることができる。
%
% $C_i$は$L_i$の上流で分岐したレイヤー ID
% の集合である。$A=\{a_1,a_2,\ndots\}$をレイヤーIDの集合としたときに、各レイヤーをどのスキー
% マにするかの組み合わせをタプル$\lambda_A=(\lambda_{a_1},\lambda_{a_2},\ndots)$で表現する。
% すなわちこのタプルの全集合は$A$に関する$\Lambda$のカーテジアンパワーセットであ
% る。$\lambda_A$が制約条件にあるとき、すべての$\lambda_j~(j\in A)$に対
% し、$F_j(\lambda_j)\in S_i(\lambda_i;\lambda_A)$を満たさなければならないものとする。

We then examine networks where a layer output is branched into two layers, as shown
in Figure~\ref{fig:network_examples}'s net B.  By extending the idea for the straight
case, we can calculate the optimal routine subpath $S_5(\lambda_5)$ at layer 5 as the
union of $S_3(\lambda_3)$ and $S_4(\lambda_4)$ with adapt routines connecting layers 3 and
4 to layer 5.  However, this computation requires that schemas for layer 1 in
$S_3(\lambda_3)$ and $S_4(\lambda_4)$ are identical; otherwise $S_5(\lambda_5)$ may have
two routines for layer 1 and is invalid.  To handle such cases, we modify the definition
of $S_i(\lambda)$ to include a constraint $C_i$ as $S_i(\lambda; C_i)$.  For example,
$S_5(\lambda; \lambda_1 = \alpha)$ means that the routine for layer 1 must have schema
$\alpha$.

With this extension, the optimal routine subpath at layer 5 can be obtained incrementally
by the following equation:
\begin{equation}
\label{eq:recurrence_formula_constrained}
\begin{array}{l}
S_5(\alpha; \lambda_1=\alpha)=\{F_5(\alpha)\} \cup \fastest \\
\left\{\begin{array}{l}
S_3(\alpha; \lambda_1=\alpha) \cup S_4(\alpha; \lambda_1=\alpha), \\
S_3(\alpha; \lambda_1=\alpha) \cup S_4(\beta; \lambda_1=\alpha) \cup \{F^a_{4,5}(\beta \rightarrow \alpha)\}, \\
S_3(\beta; \lambda_1=\alpha) \cup \{F^a_{3,5}(\beta \rightarrow \alpha)\} \cup S_4(\alpha; \lambda_1=\alpha), \\
S_3(\beta; \lambda_1=\alpha) \cup \{F^a_{3,5}(\beta \rightarrow \alpha)\} \cup \\
~~~~S_4(\beta; \lambda_1=\alpha) \cup \{F^a_{4,5}(\beta \rightarrow \alpha)\}
\end{array}
\right.
\end{array}
\end{equation}
The $\fastest$ term considers union of the optimal routine subpaths at layers 3 and 4
(i.e., $S_3$ and $S_4$).  For each subpath, there are two candidates depending on the
selected schema, as in the straight case.  Thus the $\fastest$ term evaluates the four
candidates in total.  Similar relations for $S_5$ exist for different combinations of the
schema and constraint.

More generally, the constraint can be regarded as an element of the $m$th Cartesian power
of routine schemas $\Lambda$ when there are $m$ constraint layers.  The $m$th Cartesian
power of $\Lambda$ is
\begin{equation}
  \Lambda^m \equiv \underbrace{\Lambda \times \cdots \times \Lambda}_m.
\end{equation}
When we let $A = \{a_1, \ldots \, a_m\}$ be a set of indices of $m$ constrained layers, a
tuple $\lambda_A=(\lambda_{a_1}, \ldots, \lambda_{a_m}) \in \Lambda^m$ represents a
combination of schema constraints for the layers.  When $\lambda_A$ is in the constraint,
the fastest routine for layer $j$ with schema $\lambda_j$, $F_j(\lambda_j)$, must be
included in the constrained optimal routine path, $S_i(\lambda_i; \lambda_A)$.  Based on
this notation, we can show the following proposition.

\begin{proposition}[Optimality of constrained DP]
%拘束条件付き、最適ルーチンサブパスから以下のようにして求めた$\hat{S}$は$S^*$に等しい。
  The routine path $\hat{S}$ obtained by the constrained optimal routine subpath as
  follows is optimal.
  \begin{equation}
    \hat{S}=\fastest_{\lambda\in\Lambda,\lambda_{C_n}\in\Lambda^{|C_n|}} S_N(\lambda;\lambda_{C_n}).
  \end{equation}
\end{proposition}
%\begin{proof}
  % 証明:拘束条件付き DP 変数の定義と、分岐をマージする際の漸化
  % 式\ref{eq:recurrence_formula_constraint}$を一般化することにより導かれる。
\noindent{\textit{Proof.}  Derived by the definition of constrained DP variables and generalizing the recurrence
  formula (Equation \ref{eq:recurrence_formula_constrained}) in merging branches.}
%\end{proof}

\subsection{Multi-branching Net}
%図\ref{fig:network_examples}Cのように多重の分岐を含む場合での最適化について考える。
% 分岐ネットワークの場合に従って、レイヤー１以降では拘束条件にレイヤー１のスキーマ（x2）が加わり、レイヤー２以降では拘束条件にレイヤー２のスキーマ（x2）が加わる。
% レイヤー２以降のレイヤー３，４，５，８については４通りの拘束条件と、自分自身のスキーマの２パターンについて、$S_i(\lambda\mid C)$を計算する。
% 一見すると分岐の個数が$b$になると、下流のDPで考慮する組み合わせ数が$O(2^b)$となってしまう。
% しかし、以下のように分岐のマージ点で拘束条件を緩和することがそれを防ぐ。
% レイヤー8で、レイヤー5とレイヤー７の最適ルーチンサブパスをマージするときに必要な拘束条件は、レイヤー１のスキーマだけである.
% レイヤー２のスキーマの拘束条件は、レイヤー５のマージをするときに以下のように緩和することができる。
Now we inspect networks with multiple branches illustrated by the net C in Figure
\ref{fig:network_examples}.  As a naive extension, we could apply the idea for branching
nets by constraining each branch.  For instance, layers 1 and 2 need to be constrained in
the example.  In this case, we need to evaluate eight patterns in total to compute the
optimal routine subpaths of layers at the downstream of layer 2 (i.e., layers 3, 4, 5, and
8), because the unconstrained layers also have two schema options.  In short, a naive
extension of the branching case results in $2^b$ combinations of schemas for nets with $b$
branches.

To alleviate this cost, we relax the constraint when merging branches.  For example, when
merging the subpaths of layers 5 and 7 for layer 8, the necessary constraint is only the
schema of layer 1.  This is because the constraint on layer 2 must be already satisfied at
layer 5, when merging layers 3 and 4.  Thus, the constraint on layer 2 can be relaxed at
layer 5 as follows:
\begin{equation}
S_5(\alpha;\lambda_1=\alpha)=\fastest \left\{
\begin{array}{l}
S_5(\alpha;\lambda_1=\alpha,\lambda_2=\alpha),\\
S_5(\alpha;\lambda_1=\alpha,\lambda_2=\beta).
\end{array}
\right.
\end{equation}
In this equation, we remove the constraint of $\lambda_2$ by selecting the faster subpath.
Similar relaxation can also be applied for $S_5(\alpha;\lambda_1=\beta)$,
$S_5(\beta;\lambda_1=\alpha)$, and $S_5(\beta;\lambda_1=\beta)$.  A more general condition
for the relaxation is summarized as the following proposition.

\begin{proposition}[Condition for relaxing constraints]
  \label{prop:mergeability}
  % 分岐レイヤー$L_i$のすべての下流パスが通過するレイヤーの集合$A$を考える。$L_j$におい
  % て$L_i$に関する拘束条件を緩和することができる必要十分条件は$L_j\in A$である。
  Let $A_i$ be a set of layers through which all downstream paths of branching layer $L_i$
  pass. The constraint on $L_i$ at $L_j$ can be relaxed if and only if $L_j\in A_i$.
\end{proposition}
%\begin{proof}
%証明：$L_j \in A$であれば$L_i$のすべての下流パスが$L_j$を通る。
% $L_j$より下流パスで$L_i$についての拘束条件が必要なマージが生じないので、$L_j$で拘束条件を緩和することができる。
% 逆に$L_j \not\in A$ならば、$L_i$を通り$L_j$を通らないパス$\rho$がある。
% いま定義\ref{def:routine_path}によりNetの出力は１個なので、$L_k (j < k\leq N)$で$\rho$と$L_j$の下流パスは必ず交差する。
% このとき$L_i$における拘束条件が必要になるので、$L_j$で拘束条件を緩和することはできない。
\noindent{\textit{Proof.}  If $L_j \in A_i$, then all downstream paths of $L_i$ go through $L_j$.  At the
  downstream of $L_j$, merging with a constraint on $L_i$ does not occur. % Why?
  Hence we can relax the constraint on $L_j$ in this case.}

  Conversely, if $L_j \not\in A_i$, then there is a path $\rho$ that passes through $L_i$
  without travelling $L_j$.  Besides, there exists a path $\rho'$ at the downstream of
  $L_j$ that intersects $\rho$ at $L_k (j < k \leq n)$, because a net has only one output.
  Since the constraint on $L_i$ is required at the intersecting point, we cannot relax the
  constraint at $L_j$.
%\end{proof}

\begin{algorithm}[!t]
\caption{DPRS for Hybrid Tuning}
\label{alg:hybrid_tuning}
\begin{algorithmic}[1]
  \STATE \textbf{Input:} 1-in-1-out net $\{L_i\}$, Layer-wise optimal routines $\{F_i\}, \{F^a_{i,j}\}$, Routine schema set $\Lambda$
  \STATE \textbf{Output:} Optimal routine path $S^*$
  \STATE $C_0\gets \emptyset$
  \FOR {$\lambda \in \Lambda$}
  \STATE $S_0(\lambda ; \emptyset) \gets \emptyset$
  \ENDFOR
  \FOR {$i$ in $1, \ldots, N$}

  \STATE $H\gets $ Indices of layers directly connected to $L_i$
  \STATE $C_i \gets C_h$ from one of $h\in H$ (All $C_h$ are equal) % ???
  \FOR {$\lambda \in \Lambda$}
  \FOR {$\lambda_{C_i} \in \Lambda^{|C_i|}$}
  \STATE $S_i(\lambda;\lambda_{C_i}) \gets \{F_i(\lambda)\}\cup $ \\ $~~\fastest\limits_{\lambda_H\in\Lambda^{|H|}} \bigcup\limits_{j\in H} S_j(\lambda_j;\lambda_{C_i}) \cup \{F^a_{j,i}(\lambda_j\rightarrow \lambda)\}$
  \ENDFOR
  \ENDFOR
  \IF {$L_i$ is a merging layer}
  \STATE $K\gets$ Subset of $C_i$ to be relaxed at $L_i$
  \STATE $C_i\gets C_i \backslash K$
  \FOR {$\lambda \in \Lambda$}
  \FOR {$\lambda_{C_i}\in\Lambda^{|C_i|}$}
  \STATE $S_i(\lambda;\lambda_{C_i})\gets \fastest\limits_{\lambda_K\in \Lambda^{|K|}}
  S_i(\lambda;\lambda_{C_i},\lambda_K)$
  \ENDFOR
  \ENDFOR
  \ENDIF

  \IF {$L_i$ is a branching layer}
  \FOR {$\lambda\in\Lambda$}
  \FOR {$\lambda_{C_i}\in\Lambda^{|C_i|}$}
  \STATE $S_i(\lambda;\lambda_{C_i},\lambda_i=\lambda)\gets S_i(\lambda;\lambda_{C_i})$
  \ENDFOR
  \ENDFOR
  \STATE $C_i\gets C_i \cup \{i\}$
  \ENDIF

  \ENDFOR
  \STATE $S^* \gets \fastest\limits_{\lambda\in\Lambda} S_N(\lambda;\emptyset)$
\end{algorithmic}
\end{algorithm}

\begin{table*}[t]
  \begin{center}
    \renewcommand{\arraystretch}{1.2}
    \caption{Average inference times in milliseconds with standard deviations in
      parentheses, under various tuning options.}
    \label{table:hybrid_tuning_experiments}
    \begin{tabular}{@{}lcrrrr@{}}
      \toprule
      && \multicolumn{4}{c}{Tuning option} \\ \cmidrule{3-6}
      Network && \multicolumn{1}{c}{float32 (w/o tune)} & \multicolumn{1}{c}{float32} & \multicolumn{1}{c}{qint8} & \multicolumn{1}{c}{hybrid} \\
      \midrule
      VGG16 && 548.442 ($\pm{16.904}$) & 219.603 ($\pm{0.963}$) & 372.658 ($\pm{1.251}$) &
                                                                                    198.453 ($\pm{3.758}$) \\
      ResNet50 && 168.179 ($\pm{0.094}$) & 111.344 ($\pm{1.606}$) & 119.607 ($\pm{0.527}$) &
                                                                                      102.544 ($\pm{1.373}$) \\
      MobileNetV2 && 20.295 ($\pm{0.044}$) & 16.246 ($\pm{0.056}$) & 15.409 ($\pm{0.017}$) &
                                                                                      14.931 ($\pm{0.019}$) \\
      MobileNetV3 && 7.832 ($\pm{0.034}$) & 5.722 ($\pm{0.038}$) & 9.142 ($\pm{0.040}$) & 5.695 ($\pm{0.066}$) \\
      \bottomrule
    \end{tabular}
  \end{center}
\end{table*}

\begin{table*}[tb]
  \begin{center}
    \renewcommand{\arraystretch}{1.2}
    \caption{Average inference times of inference frameworks, shown in milliseconds with
      standard deviations in parentheses.  TensorFlow Lite and PyTorch Mobile models were
      quantized to 8-bit. \Softneuro{} used hybrid tuning of float32 and qint8.}
    \label{table:framework_comparison}
    \begin{tabular}{@{}lrrr@{}}
      \toprule
      Network & \multicolumn{1}{c}{TensorFlow Lite} & \multicolumn{1}{c}{PyTorch Mobile} & \multicolumn{1}{c}{Ours} \\
      \midrule
      VGG16 & 403.062 ($\pm{12.053}$) & N/A & 198.453 ($\pm{3.758}$) \\
      ResNet50 & 129.698 ($\pm{0.871}$) & 152.172 ($\pm{0.766}$) & 102.544 ($\pm{1.373}$) \\
      MobileNetV2 & 25.730 ($\pm{0.639}$) & 55.653 ($\pm{0.031}$) & 14.931 ($\pm{0.019}$) \\
      MobileNetV3 & 7.479 ($\pm{0.036}$) & N/A & 5.695 ($\pm{0.066}$)\\
      \bottomrule
    \end{tabular}
  \end{center}
\end{table*}

\subsection{Multi-input, Multi-output Net}
\label{sec:mimo_net}

%これまではDNNの入出力層が１個ずつの場合を考えていた。
% 近年では複数の入力を取る画像の合成処理や、複数の出力を出すマルチタスク学習用のDNNなどが成功を収めている。
% そのような場合でも、ルーチンパスを定義し、上記の議論を適用するために、すべての入力レイヤーに接続する補助入力レイヤーと、すべての出力レイヤーが接続する補助出力レイヤーを導入する（図\ref{fig:network_examples}D。
% 補助入出力レイヤーは処理時間がゼロの単一のルーチンを持ち、前後レイヤーとのadaptレイヤーによる接続を必要としない。

So far, we have explored only 1-in-1-out networks, but networks with multiple inputs and
outputs have recently been utilized for tasks such as synthesizing images and multi-task
learning~\cite{gatys:cvpr16, vandenhende:tpami21}.  To adapt the above discussion to such
cases as well, we introduce \emph{auxiliary input and output layers} as shown in the net D
(Figure \ref{fig:network_examples}).  The auxiliary input layer connects to all input
layers, and the auxiliary output layer merges all output layers.  The auxiliary layers
have a single routine with zero processing time, so there is no impact on performance
optimization.

\subsection{Algorithm}
\label{sec:dprs_algorithm}
%以上の議論を一般化したHybrid Tuningはアルゴリズム\ref{alg:hybrid_tuning}のような疑似コードで表される。
%複数入出力を持つDNNは1-in-1-outのDNNに変換し、レイヤーをトポロジカルオーダーで再インデクスしたのちにアルゴリズムを適用する。
% また\mathrm{Net}がチャイルドネットを持つ場合には、最も内側のチャイルドネットから再帰的にアルゴリズムを適用する。
% 簡単のために、式(\ref{eq:power_set})では常にadaptルーチンをルーチンパスに加えているが、必要ない場合はvoid adaptルーチン（$T(r^a_{i,j})=0$）を挿入すると考える。
% 19行目で$S^*$を求めるときは、拘束条件が必ず空集合になることに注意する。これは命題\ref{prop:mergeability}より明らかである。
Algorithm \ref{alg:hybrid_tuning} shows the pseudo-code for hybrid tuning, a
generalization of the above discussion.  The algorithm receives a 1-in-1-out net; thus a
network with multiple inputs and outputs is converted, with layers re-indexed in
topological order, before applying the algorithm.  Meanwhile, if the input net has a
childnet, the algorithm is applied recursively starting from the innermost childnet.  For
simplicity, we always add adapt routines at line 12, and insert a void adapt routine
($T(r^a_{i,j})=0$) when it is not needed.  Note that the constraint when obtaining $S^*$
at line 33 is always an empty set, because at the output layer, all constraints should
already be satisfied.

\begin{table}[tb]
  \begin{center}
    \renewcommand{\arraystretch}{1.2}
    \caption{Selected routine descriptors and parameters for VGG16 by hybrid tuning.}
    \label{table:selected_routines}
    \begin{tabular}{@{}lll@{}}
      \toprule
      Layer & Routine Descriptor & Routine Parameters \\
      \midrule
      block1 conv1 & cpu/owc32\_neon & cache:8192 \\
       & & task\_ops:32768 \\
      block1 conv2 & cpu/wg2 & tile\_size:4\\
      block1 pool & cpu/neon\\
      block2 conv1 & cpu/wg2 & tile\_size:8\\
      block2 conv2 & cpu/wg2 & tile\_size:6\\
      block2 pool & cpu/neon\\
      block3 conv1 & cpu/wg2 & tile\_size:8\\
      block3 conv2 & cpu/wg2 & tile\_size:8\\
      block3 conv3 & cpu/wg2 & tile\_size:8\\
      block3 pool & cpu/neon\\
      block4 conv1 & cpu/wg2 & tile\_size:8\\
      block4 conv2 & cpu/wg2 & tile\_size:4\\
      block4 conv3 & cpu/wg2 & tile\_size:4\\
      block4 pool & cpu:qint8/neon \\
      block5 conv1 & cpu/wg2 & tile\_size:6\\
      block5 conv2 & cpu/wg2 & tile\_size:6\\
      block5 conv3 & cpu/wg2 & tile\_size:6\\
      block5 pool & cpu:qint8/neon \\
      flatten & cpu:qint8 \\
      fc1 & cpu:qint8 \\
      fc2 & cpu:qint8 \\
      fc3 & cpu:qint8 \\
      softmax & cpu/naive \\
      \bottomrule
    \end{tabular}
  \end{center}
\end{table}

\section{Experiments}
\label{experiments}

We evaluate inference performance on the VGG16 \cite{simonyan2015very}, ResNet50
\cite{he2016deep}, MobileNetV2 \cite{sandler2018mobilenetv2} and MobileNetV3
\cite{howard2019searching} models, which are commonly used for classification or as
backbones for other tasks.  Inference performance was measured on a Samsung Galaxy S8 containing a Snapdragon 835 SOC
\cite{snapdragon835}.  We run inference 20 times per experiment for
obtaining the averages and standard deviations.  We also compare the tuning and inference speeds of
\softneuro{} with those of TVM.

%VGG16, resnet50, mobilenet あたりでチューニング無し、ありの比較.
%プロファイル時間がどれくらいかかるかの話.

%float32だけ, qint8だけ, float32+qint8でのハイブリッドチューニングの比較.

\subsection{Hybrid Tuning Performance}
\label{Effect_of_the_Hybrid_Tuning}

We compare the inference speeds of four cases: (1) without tuning, (2) tuning only for
float32 routines, (3) tuning only for qint8 routines, and (4) hybrid tuning of float32 and
qint8.  The result is summarized in Table~\ref{table:hybrid_tuning_experiments}.

Hybrid tuning achieves the best result for all tested models, by as much as 9.6\%
compared to float32-only tuning on VGG16.  This is because it considers not only the
inference speed of each routine, but also the conversion costs between float32 and qint8,
allowing for the optimal combination of float32 and qint8 routines.  It is of note that
float32-only tuning achieved better results than qint8, except for MobileNetV2.  This
reveals that float32 routines can be faster than qint8 routines in some conditions, in
this case, due to Winograd's efficiency for float32.  In fact, qint8 routines were selected
for only six layers of VGG16, as shown in Table~\ref{table:selected_routines}.
Specifically, the qint8 routines are used for \texttt{block4 pooling}, \texttt{block5
  pooling}, \texttt{flatten}, \texttt{fc1}, \texttt{fc2}, and \texttt{fc3}.  Note that
\texttt{block4 pooling} is situated between float32 routines.  Such routine selection is
enabled by hybrid tuning; the tuning algorithm determines that the qint8 routine is faster
than float32 routines even if quantization and dequantization need to be performed.

%他の推論フレームワーク (SNPE, OpenVINO, Tensorflow Lite, TVM, ...) との比較.

\subsection{Comparison with Other Frameworks}
\label{Comparison_with_other_frameworks}

We compare \softneuro{} inference speeds with two inference frameworks: TensorFlow Lite
(ver. r2.4) and PyTorch Mobile (ver. 1.7.1).  Table~\ref{table:framework_comparison} shows
the result of experiments.  PyTorch Mobile does not have readily available VGG16 and
MobileNetV3 models, so they were not measured.  For a fair comparison with hybrid tuning,
quantized models were used in TensorFlow Lite and PyTorch Mobile.

\Softneuro{} outperforms TensorFlow Lite and PyTorch Mobile for all models.
Notably, \softneuro{} is more than 2 times faster than TensorFlow Lite on VGG16. Compared with
PyTorch Mobile, \softneuro{} is more than 3 times faster on MobileNetV2.

% how much better?
% more frameworks

\subsection{Tuning Speed}
\label{tuning_time}

\begin{table}[tb]
  \begin{center}
    \renewcommand{\arraystretch}{1.2}
    \caption{Tuning and average inference times.}
    \label{table:tuning_time}
    \begin{tabular}{@{}lrrcrr@{}}
      \toprule
      \multicolumn{1}{l}{}  & \multicolumn{2}{c}{Tuning Time (s)} && \multicolumn{2}{c}{Inference Time (ms)} \\ \cmidrule{2-3} \cmidrule{5-6}
      Network & \multicolumn{1}{c}{TVM} & \multicolumn{1}{c}{Ours} && \multicolumn{1}{c}{TVM} & \multicolumn{1}{c}{Ours} \\
      \midrule
      VGG16  & 33,240 & 66 && 84.928 & 18.498 \\
      ResNet50  & 127,900 & 14 && 23.881 & 11.901\\
      \bottomrule
    \end{tabular}
  \end{center}
\end{table}

We measured the tuning and inference speeds for VGG16 and ResNet50
using TVM and \softneuro{} on an Intel Xeon Gold 6126 CPU (Table ~\ref{table:tuning_time}).
TVM was set to quantize models to 8-bit and tune with 2,000 trials per operation, matching
the configuration used in the TVM paper~\cite{tvm}.
\Softneuro{} shows more than an order of magnitude faster tuning over TVM while also achieving as
much as thrice faster inference speeds.
\section{Conclusions and Future Work}
\label{conclusions.tex}

% 本論文で我々はパフォーマンス最適化のための新しい推論フレームワークを提
% 案した. 基本的なアイデアはルーチンをレイヤーから分離したことである. ルー
% チンはあるレイヤに対しての計算の実体であり, あるレイヤに対して同じ計算
% 結果を出力する複数のルーチンがありうる.
% 本フレームワークでは柔軟なパフォーマンス最適化のために実行環境上である
% レイヤに対するあらゆるルーチンを実行(プロファイリング)する.
% 最適化は各レイヤの最速ルーチンを伝達ブロブの変換コストも考慮して求める
% という問題に定式化され,それは動的計画ルーチン選択法により解くことがで
% きる.
% 我々はチャイルドネットを導入し, これによりレイヤを組み合わせてあるルー
% チンを実装することを可能にし効率良い最適化を実現している.

% ------------------------------------------------------------------
% Basic Architectureの章へ移動
% ------------------------------------------------------------------
% プロファイリングはレイヤ毎の単体プロファイリングと, その後の確率的統合
% プロファイリングの2つのフェーズに分けられる
% これは単体プロファイリングはそのキャッシュヒット率が実際の実行時と異な
% りその測定処理時間が実際のものとは異なるかもしれないからである.
% 最終的な最適化はこの確率敵統合プロファイリングの結果を元に, 動的計画ルー
% チン選択法により導出される.
% Profiling is divided into two phases: unit profiling at each layer
% and then integrated profiling.
% This is because the cache hit rate differs between unit profiling and
% actual execution, affecting processing time measurements.
% The final optimization is performed by DPRS based on the result of
% integrated profiling.

% 我々の実験は本フレームワークは優秀であり, 他の推論フレームワークを超え
% るパフォーマンスを達成することを示している.
% また本フレームワークは他の高速化手法, レイヤフュージョン, float16演算,
% 量子化, Winograd, レイヤフュージョなどとも組み合わせることができるので
% 非常に効果的である.

In this paper, we have described a novel, performance-optimized inference framework.
The core idea is to separate routines from layers, thereby enabling efficient and flexible
performance optimization.  The optimization has been formulated as a problem of finding
the fastest routine at each layer while taking into account the adaptation cost.  To
efficiently solve the problem, we have proposed the DPRS algorithm based on dynamic
programming.  Our experiments show that the proposed framework achieves faster inference
and more efficient tuning than other frameworks.

% ------------------------------------------------------------------
% 分岐ネスト問題は省略
% ------------------------------------------------------------------
% \begin{figure}[ht]
%   \begin{center}
%   \subfigure[ResNet-style skip connections]{
%     \includegraphics[width=2.5cm]{conclusions_resnet.eps}
%   }
%   \hspace{5mm}
%   \subfigure[Unet-style nested branches]{
%     \includegraphics[width=2.5cm]{conclusions_unet.eps}
%   }
%   \caption{Branch patterns.}
%   \label{conclusions-resnet-unet}
%   \end{center}
% \end{figure}
% 動的計画ルーチン選択法では, 各分岐はルーチンスキーマの組み合わせ数を増や
%し, 各マージはその組み合わせ数は減少させる.
% Resnetのようなスキップコネクションを持つモデルでは分岐は直ちにマージさ
% れその組み合わせ数は限定的である.
% 一方でUnetのように分岐がマージせずにネストし, そのネスト分岐が非常に深
% いモデルでは組み合わせ爆発が起こるかもしれない.
% 我々はそのような深いネストを持つモデルにまだ直面したことはないが, これ
% は将来的に課題になるかもしれない.
% In DPRS, each branch increases the number of routine schema combinations,
% and each merge decreases the number of combinations.
% In a model with skip connections such as ResNet
% (Figure~\ref{conclusions-resnet-unet}~(a)), braches are immediately
% merged and the number of combinations is limited.
% On the other hand, in a model where branches are nested without merging
% such as Unet (Figure~\ref{conclusions-resnet-unet}~(b)) and the nested
% branches are very deep, a combinatorial explosion may occour.
% We have not yet faced a model with such deep nesting, but this may be a
% problem in the future.

% 本フレームワークは現在, 処理時間に対してのみ最適化を実行しており、そこ
% では量子化手法が許容範囲を超えて精度を劣化させる可能性があり, Winograd
% アルゴリズムもタイルサイズに応じてカーネルサイズが大きくなるため大量の
% メモリを消費する可能性がある.
% そのような精度劣化やメモリ消費量を制限した上で, その処理時間の最適化を
% 行うということは将来課題の1つである.

In future we plan to optimize the processing time with restrictions of accuracy and
memory usage.  \Softneuro{} currently considers only processing time when tuning the
performance.  However, quantization may degrade accuracy beyond acceptable limits.
Meanwhile, the Winograd algorithm may consume a large amount of memory due to kernel sizes
that increase with tile sizes.  Such accuracy degradation and excessive memory consumption
should be alleviated in the future.

%\section{Prepare Your Paper Before Styling}
%\subsection{Figures and Tables}
%\paragraph{Positioning Figures and Tables} Place figures and tables at the top and 
%bottom of columns. Avoid placing them in the middle of columns. Large 
%figures and tables may span across both columns. Figure captions should be 
%below the figures; table heads should appear above the tables. Insert 
%figures and tables after they are cited in the text. Use the abbreviation 
%``Fig.~\ref{fig}'', even at the beginning of a sentence.

%\begin{figure}[htbp]
%\centerline{\includegraphics{fig1.png}}
%\caption{Example of a figure caption.}
%\label{fig}
%\end{figure}

%Figure Labels: Use 8 point Times New Roman for Figure labels. Use words 
%rather than symbols or abbreviations when writing Figure axis labels to 
%avoid confusing the reader. As an example, write the quantity 
%``Magnetization'', or ``Magnetization, M'', not just ``M''. If including 
%units in the label, present them within parentheses. Do not label axes only 
%with units. In the example, write ``Magnetization (A/m)'' or ``Magnetization 
%\{A[m(1)]\}'', not just ``A/m''. Do not label axes with a ratio of 
%quantities and units. For example, write ``Temperature (K)'', not 
%``Temperature/K''.

%\vspace{12pt}
%\color{red}
%IEEE conference templates contain guidance text for composing and formatting conference papers. Please ensure that all template text is removed from your conference paper prior to submission to the conference. Failure to remove the template text from your paper may result in your paper not being published.

\end{document}